\documentclass{article}

\usepackage[final]{nips_2016}

\usepackage[utf8]{inputenc} 
\usepackage[T1]{fontenc}    
\usepackage{hyperref}       
\usepackage{url}            
\usepackage{booktabs}       
\usepackage{amsfonts}       
\usepackage{nicefrac}       
\usepackage{microtype}      
\usepackage{graphicx}
\usepackage{float}

\title{End-to-End Visual Odometry with RNNs and Attention}

\author{
  Ruiyu Li \\
  \texttt{ruiyul@cs.cmu.edu}
   \And
  Yinjia Liu \\
  \texttt{yinjial@cs.cmu.edu}
   \And
  Alexander Yu \\
  \texttt{ayu2@cs.cmu.edu}
}

\begin{document}

\maketitle

\section{Introduction}

    Video Odometry (VO) is the process of estimating the ego-motion of an object by analyzing visual information such as a sequence of frames from one or multiple cameras. It has been a popular research topic in computer vision and robotics, and its applications include mobile robotic systems as well as autonomous driving. In this project, we investigate existing end-to-end deep-learning approaches to VO, and propose a novel temporal attention-based model to improve upon the baseline. In addition, while the vast majority of existing deep-learning-based approaches to VO are trained on driving data, we investigate the performance of deep-learning-based VO to the more dynamic and complex problem of hand-held cameras.

\section{Background}
    Our preliminary work consists of implementing the DeepVO Recurrent CNN \cite{DeepVO} that appends a RNN with two LSTM layers to a CNN as a baseline model. Despite the promising results of DeepVO on the KITTI VO Benchmark \cite{KITTI}, this baseline we implemented did not perform well on the TUM RGB-D dataset. The training loss achieves a minimum value of 0.027, corresponding to a root-mean-square (RMS) deviation between the true and predicted trajectories of 0.23m. The validation loss, however, fails to converge and fluctuates about a mean value of 3.91, corresponding to a RMS deviation of 2.80m. 
    
\section{Related Work}

    \subsection{Feature-Based Methods}
    Feature-based methods extract features such as edges and corners in each frame, and determine motion by matching extracted feature points across different frames. However, these methods are highly sensitive to noise in the images. Several different approaches exist to address this problem. For example, visual Simultaneous Localisation and Mapping (SLAM) is utilized to minimize reprojection error through bundle adjustment. Examples of applications of visual SLAM include DOT \cite{DOT}, which performs well especially in highly dynamic scenes.
    
    \subsection{CNN-Based Methods}
    There also exist deep learning-based methods, in particular CNN-based methods. Konda and Memisevic \cite{Konda} designed a deep learning architecture to extract visual motion and depth information, and then transform the extracted representations into velocity and direction information via a Convolutional Neural Network (CNN). McFall et al. \cite{McFall} proposed another CNN that is able to predict changes in orientation of a camera as well as relative translational motion.
    
    \subsection{Recurrent CNN (RCNN)-Based Method}
    Wang et al. \cite{DeepVO} designed a new model called DeepVO that involves the use of a Recurrent Neural Network (RNN) for sequential learning, in addition to the use of a CNN for feature extraction. The CNN features are treated as input to a deep RNN which employs a Long Short-Term Memory (LSTM) that is able to learn long-term dependencies in image sequences.

\section{Methods}

    \subsection{Data and Preprocessing}
\begin{figure}
        \centering
        \includegraphics[width = 0.4\textwidth]{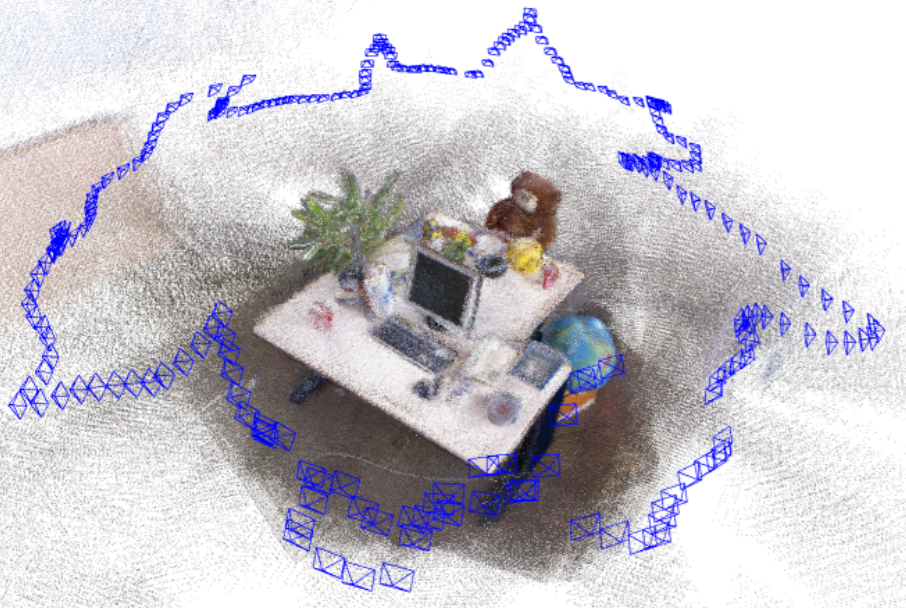}
        \caption{An example trajectory in the TUM RGB-D dataset generated from a 3D reconstruction by Mur-Atal in \cite{ORB-SLAM}}
        \label{fig:example_trajectory}
    \end{figure}

    We use the TUM RGB-D benchmark \cite{RGBD} dataset consisting of 39 sequences of color and depth images in full sensor resolution. 35 videos in the dataset are captured by a handheld Microsoft Kinect camera, while 4 video sequences are captured by a Kinect camera mounted to a wheeled robot. For all videos, the ground truth trajectory is captured by external motion capture cameras. An example trajectory is shown in Figure \ref{fig:example_trajectory}.
    
    The dataset has three separate data sources: RGB camera images, depth camera images, and ground truth trajectories. Each source is captured asynchronously. Prior to model training, we synchronize each source to the timestamps of the RGB camera images. For each RGB image, we select the depth image with the closest timestamp, and interpolate between the two closest ground truth trajectory labels. Furthermore, all images are normalized by subtracting the sample mean and dividing by the sample standard deviation for each channel.
    
    \subsection{Baseline Models}
    We use an architecture inspired by the DeepVO Recurrent CNN \cite{DeepVO} as a baseline model. While DeepVO reports excellent results on the KITTI autonomous driving dataset, its performance on the TUM RGB-D dataset is unknown. We first implement and train DeepVO as described by its authors. Then, in an attempt to adapt the original model to the more difficult task presented by the TUM RGB-D dataset, we implement and train three modified versions of the original model which investigate the effects of pre-training and the additional information provided by the depth camera. The best of these four VO models is used as a baseline against which our attention-based approach is evaluated.
    
    The original DeepVO model takes as input a sequence of 6 channel, $480 \times 640$ images where the first 3 channels correspond to the RGB image at the current time step, and the second 3 channels correspond to the RGB image at the previous time step. Notably, it does not accept any depth images as input. Inputs are passed through a 9 layer CNN with architecture and parameters defined by the feature extractor of a pre-trained FlowNet \cite{FlowNet}. During training, the parameters of the CNN are frozen. The output of the CNN is passed into two stacked LSTMs with a hidden space of 1000 nodes each. The LSTM outputs the predicted X, Y, and Z coordinate for each time step.
    
    The first modification to DeepVO, denoted Baseline Model 1, replaces the pre-trained FlowNet parameters with a random initialization, and allows the CNN parameters to be updated during training. This modification addresses the possibility that the parameters of the FlowNet feature extractor, trained on artificially rendered images with a different downstream task, may not transfer well to the erratic, motion-blurred images of a hand-held camera. 
    
    The second modification, Baseline Model 2, also randomly initializes CNN parameters and allows them to be trained. However, instead of taking 6 channel images as input, it takes an 8 channel image as input, where each group of 4 channels corresponds to a stacked RGB image and depth image. This modification addresses both the possibility that the FlowNet parameters may be inadequate and the possibility that depth images may improve model performance.
    
\begin{figure}
        \centering
        \includegraphics[width = 0.95\textwidth]{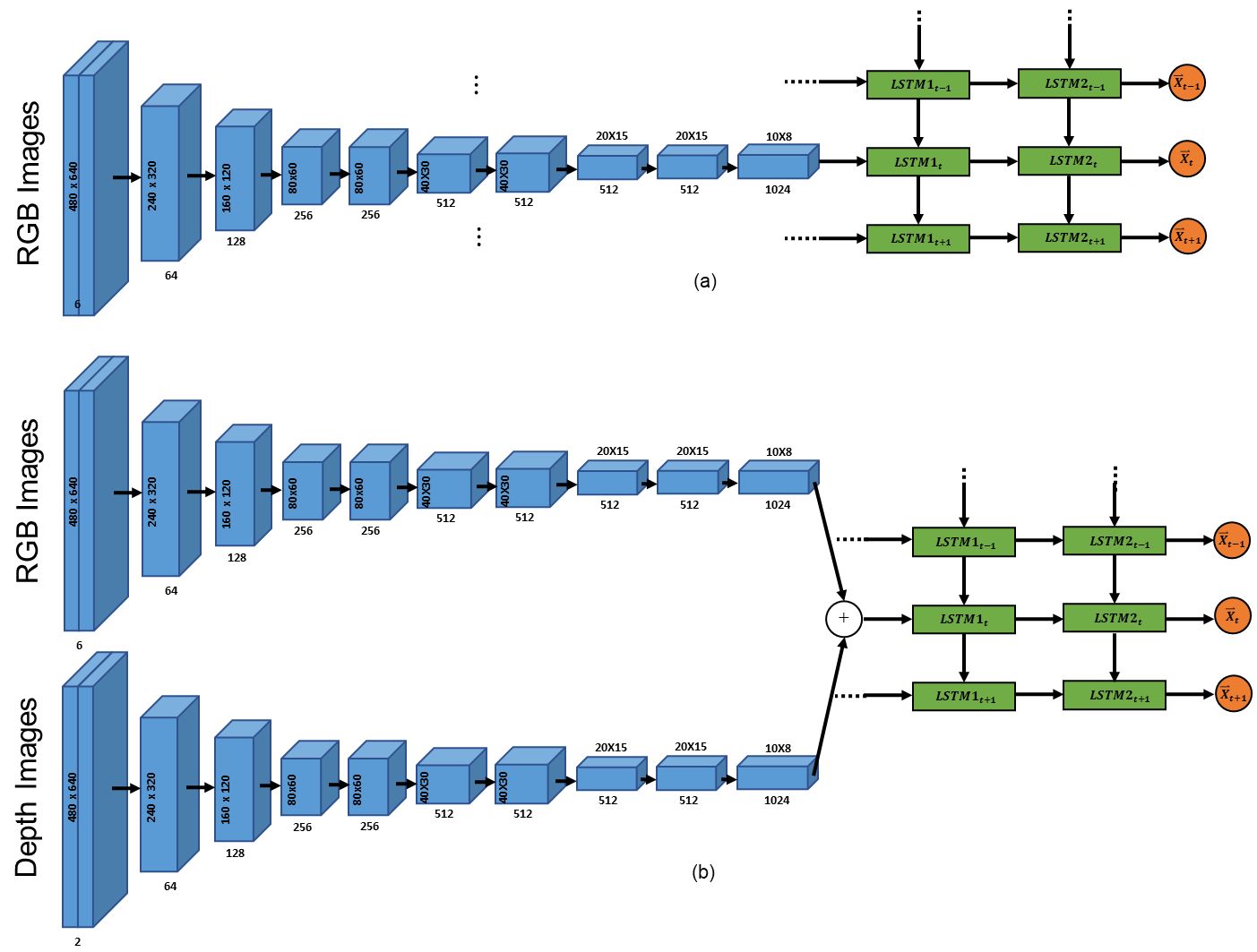}
        \caption{Model architectures of the baseline models. (a) The architecture used by the original DeepVO model, Baseline Model 1, and Baseline Model 2. The number of input channels, and trainability of convolutional layers varies between models. (b) The architecture of Baseline Model 3. The convolutional head responsible for RGB images is frozen during training, while the head responsible for depth images is trainable.}
        \label{fig:baseline_arch}
    \end{figure}

    The third modification, Baseline Model 3, preserves the architecture and parameters of the original DeepVO model. We add a second convolutional head responsible for processing the depth information in parallel with FlowNet. The outputs of this convolutional head and FlowNet are summed and passed to the LSTM. This modification addresses the possibility that the FlowNet parameters transfer well to our task, but depth images still offer improvements in performance. Figure \ref{fig:baseline_arch} illustrates the candidate baseline models' architectures.
    
    For all baseline models, training is conducted on an NVIDIA V100 Tesla for 100 epochs with the Adam optimizer and a learning rate of 1E-6. Due to the limited number of sequences, training is done with a batch size of 1. As a strong tendency to overfit was observed in preliminary tests, a dropout probability of 0.1 is applied to the LSTM layers, and a L2 weight penalty of 0.01 is applied to the parameters of the convolutional layers. We forgo a L2 penalty on the LSTM layers as Bengio et. al. indicate that L2 regularization on LSTM layers may inhibit their ability to learn long-term correlations \cite{L2_LSTM}. Loss is computed as the masked mean-squared error (MSE) between the predicted and ground truth trajectories; a mask is applied to all time steps for which ground truth data is unavailable. Each model is trained on a total of 13 image sequences and validated on 4 held-out sequences from the same dataset.
    
    \subsection{Attention-Based Models}
    In DeepVO \cite{DeepVO}, while an LSTM is used to capture long-term temporal information from a series of CNN outputs, there can still be information loss along the temporal dimension. Thus we propose attention-based models within the LSTM layers to improve utilization of long-term temporal correlations. We expect that these attention-based models can make more accurate predictions compared to the baseline while holding training and evaluation methods constant.

    Since we are using global self attention which attends to every frame in the entire sequence, it is infeasible to train the model conventionally since the entire sequence of video frames cannot fit within GPU memory. To circumvent this problem, we propose a training method that divides the training procedure into three parts.
    
    First, we divide the model into two sections: the first section consists of everything before the attention layer, and the second section consists of everything after and including the attention layer.
    
    1. We perform the forward pass on the first model section without storing gradient information to obtain the value of the input to the attention layer, which is the hidden states of the first LSTM layer (LSTM1). During this step, we divide the sequence into sub-sequences of length 64 and compute the forward pass iteratively, saving the hidden states of LSTM1 for each iteration.
    
    2. Then we perform the forward pass on the second model section with gradient information, compute the final MSE loss, and then backpropagate through the second model section. During this backpropagation, we compute the gradient of loss with respect to the input of the attention layer.
    
    3. Finally, we recompute the forward pass on the first model section with gradient information, and backpropagate the first model section with an intermediate loss function. Specifically, given the gradient of the final MSE loss with respect to the output of LSTM1, the loss function for the weights in the first model section is calculated as
    \[ \mathcal{L}_{intermediate}(w) = \frac{\partial \mathcal{L}_{MSE}}{\partial o} \cdot o\]
    where $o$ is the output of LSTM1. When differentiated, this intermediate loss function induces the conventional backpropagation equations. Since we perform the forward pass on the first model section twice in each step, we cannot allow stochasticity in the forward pass in order to use gradient information correctly. Therefore, for the attention-based models, we reduce the dropout probability to 0.

\begin{figure}
        \centering
        \includegraphics[width = 0.95\textwidth]{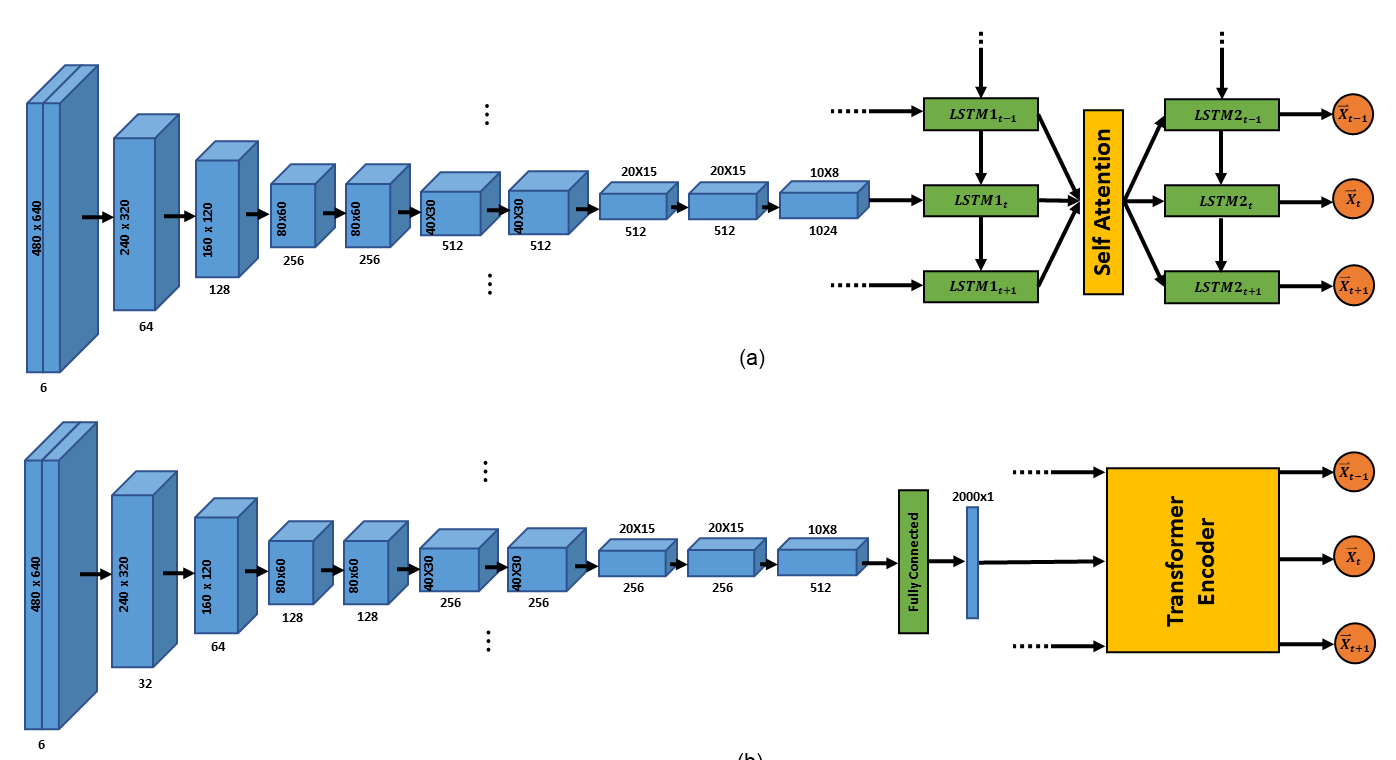}
        \caption{Architectures of the attention-based VO model. (a) The combined LSTM and attention model. (b) The transformer encoder based model. Note the reduced number of filters in the transformer based-model.}
        \label{fig:attn_arch}
    \end{figure}

    We implement two attention-based models. The first model has a single global self attention layer placed between the first LSTM layer and the second LSTM layer. The second model replaces LSTM layers entirely with a transformer encoder architecture. Since the convolutional layer maps images to an 81920 dimensional embedding, the image embeddings are intractably large as-is. Therefore for the transformer-based model, we halve the number of filters learned in each convolutional layer resulting in a 40960 dimensional embedding. We then introduce a linear layer to project this embedding to a 2000x1 vector. The transformer encoder uses 8 layers with 8 attention heads each. For both models, we do not mask attention layers, effectively allowing the model to look ahead to data in future time steps, changing the state estimation problem from a filtering problem to a smoothing problem. While we believe this may improve the models' performance, it limits the models to offline VO tasks only. Attention-based model architectures are illustrated in Figure \ref{fig:attn_arch}.
    
\section{Results}
\begin{figure}
        \centering
        \includegraphics[width = \textwidth]{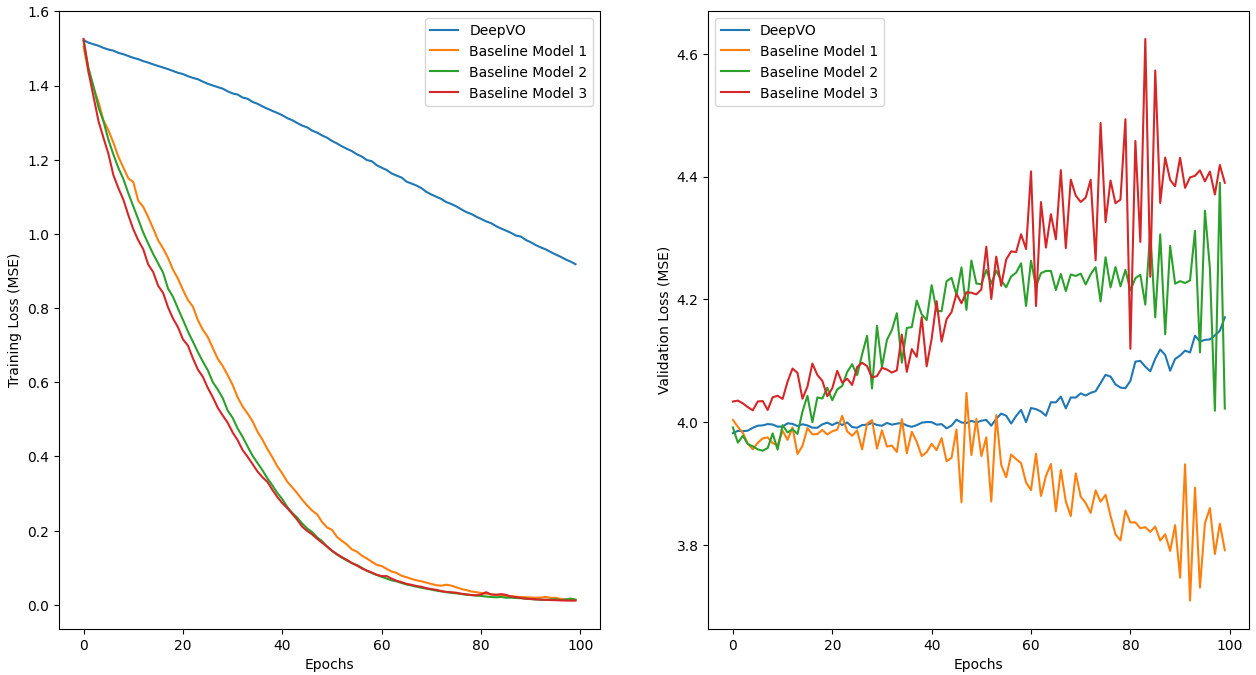}
        \caption{Training and validation losses of the original DeepVO model and the three additional baseline models on the TUM RGB-D dataset. Note the significant difference in Y-axis range between training and validation losses.}
        \label{fig:baseres}
    \end{figure}

    Figure \ref{fig:baseres} shows the training and validation losses for each baseline model. In training, the original DeepVO model is notably slower than other models to converge and fails to achieve low training loss within 100 epochs. The remaining three baseline models all achieve near-zero training loss within 100 epochs. In validation, only Baseline Model 1 is able to generalize to any extent and reduce validation loss, achieving a minimum MSE of 3.71, corresponding to an average positional error of 2.72m; the other three baseline models begin to overfit immediately and exhibit increasing validation losses with additional epochs. We will compare attention-based models to Baseline Model 1.

\begin{figure}
        \centering
        \includegraphics[width = \textwidth]{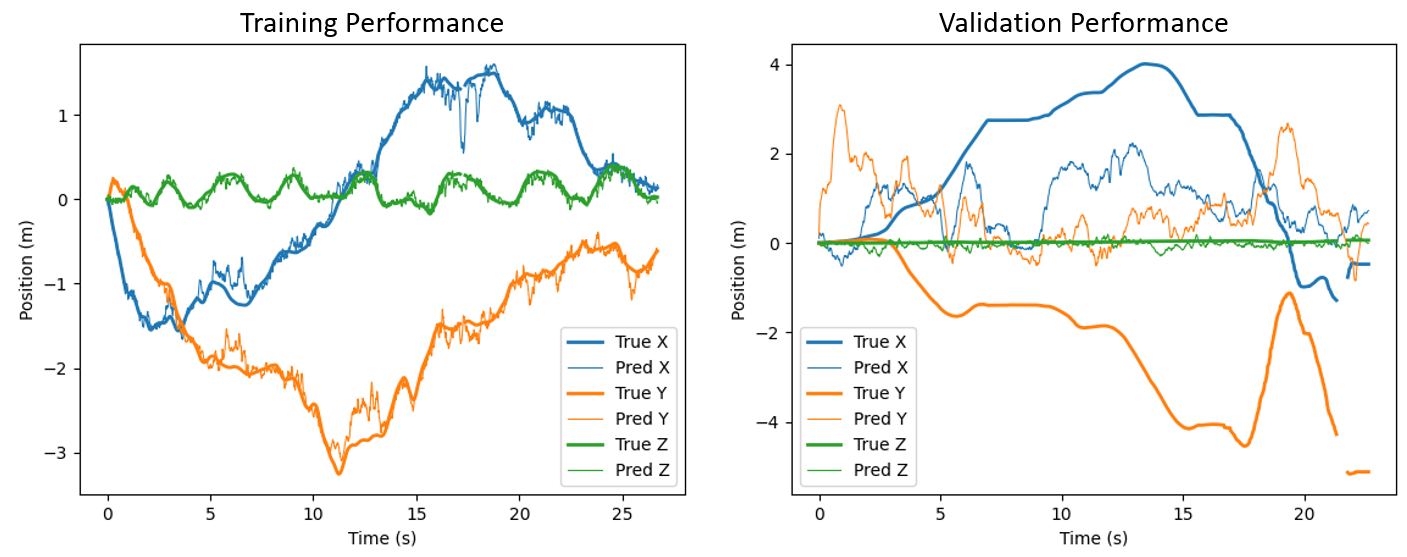}
        \caption{Representative time-series predictions generated by the best performing baseline model: Baseline Model 1. Note the approximately correct velocities predicted by the model in validation, especially from times 15s to 20s.}
        \label{fig:base_perf}
    \end{figure}

    Figure \ref{fig:base_perf} shows representative time series trajectories predicted by Baseline Model 1. In validation, we observe that even though the positional accuracy of the model is low, it is able to capture certain velocities correctly. Since velocity relies on short-term dependencies between images (differentiation at a specific time step), while position relies on long-term dependencies between images (integration over the entire sequence), we believe this indicates that the baseline model is unable to propagate information through long sequences, but is better at extracting short-term relationships between frames.
    
\begin{figure}
        \centering
        \includegraphics[width = \textwidth]{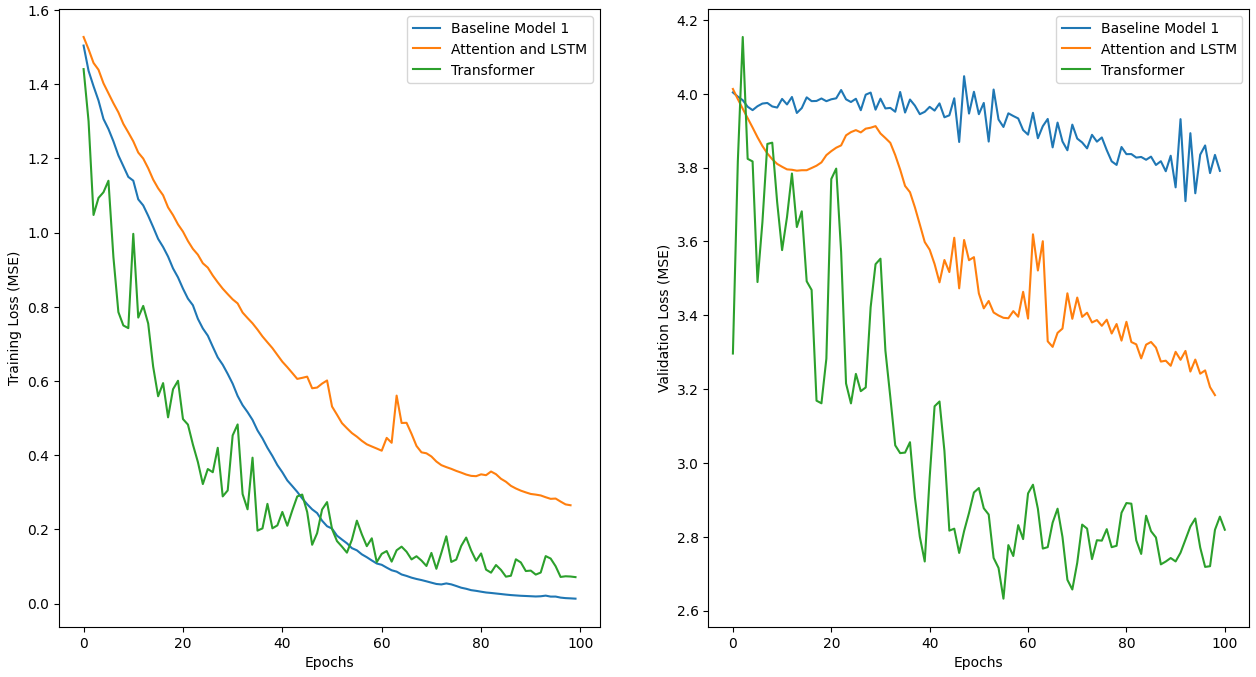}
        \caption{Training and validation losses of the attention based models compared against the best-performing baseline model.}
        \label{fig:attnres}
    \end{figure}

    Figure \ref{fig:attnres} shows the training and validation losses for each attention-based model. In training, both models are able to converge towards 0 training loss, though more slowly and erratically than the baseline models do. Nevertheless, it is clear that both attention-based models have superior generalization compared to the baseline. The attention-based model achieves a minimum MSE validation loss of 3.18, corresponding to an average positional error of 2.52m, and the transformer-based model achieves the best minimum validation loss out of all models of 2.51, corresponding to an average positional error of 2.24m. 
    
\begin{figure}
        \centering
        \includegraphics[width = \textwidth]{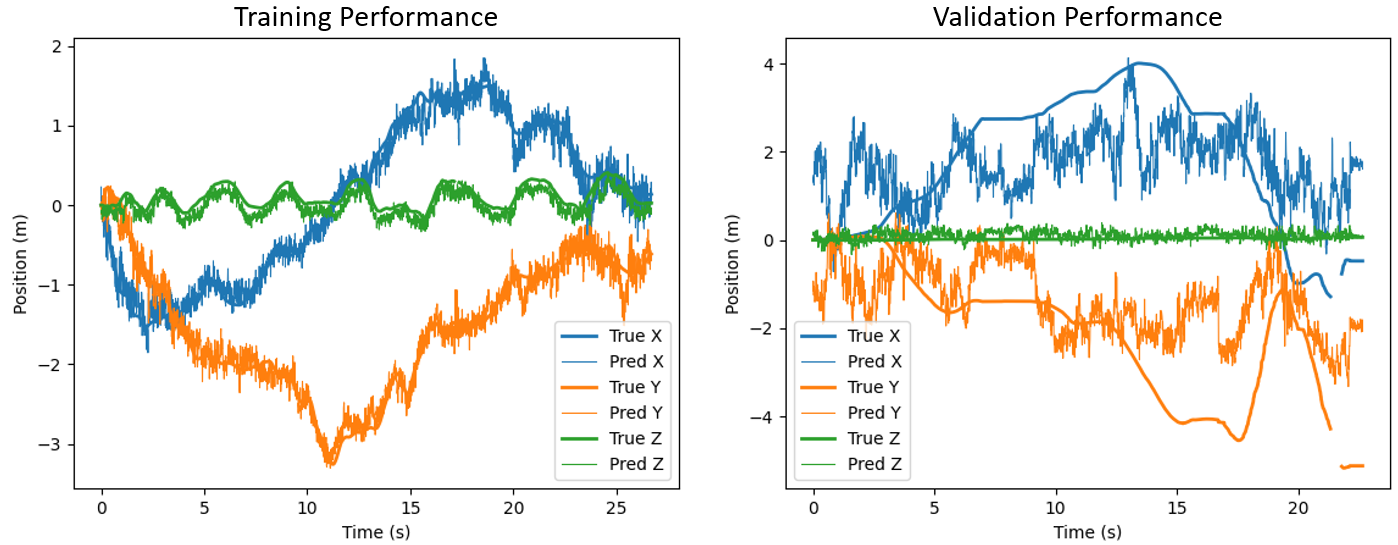}
        \caption{Representative time-series predictions generated by the transformer-based model.}
        \label{fig:trans_perf}
    \end{figure}

    Figure \ref{fig:trans_perf} shows representative time series trajectories predicted by the transformer-based model. In validation, we observe significantly improved positional accuracy when compared to baseline. However, the predicted velocities are significantly more erratic. This suggests that the transformer-based model is weaker at extracting short-term dependencies between image frames than the baseline models. We believe this may be due to the extremely long sequences, up to 5000 frames long, the model is tested on. This allows the model many opportunities to learn spurious correlations between image frames and distracts from truly important dependencies. It is possible that applying sparse attention, such as that employed by the BigBird model \cite{big_bird}, may improve performance.

\section{Discussion and Analysis}

    \subsection{Baseline models}

    As shown in the previous section, the best performing baseline model is the one that uses random initialization instead of pre-trained FlowNet parameters and ignores the depth information. We speculate that the weak performance of pretrained FlowNet parameters results from the dataset used to train FlowNet. Since FlowNet was trained on an artificially rendered dataset (specifically, 3D models of flying chairs superimposed on photo backgrounds), there is likely to be a distribution shift between that dataset and the realistic camera images of the TUM RGB-D dataset, which results in high validation losses. Regarding the cause of the weak performance of models using depth information, we suspect that the depth information in the dataset is noisy, as depth cameras often are, and results in the model learning spurious correlations within the relatively small dataset. 

    In the time series trajectory figure \ref{fig:base_perf}, the differences between the predicted trajectories of the baseline models and the true trajectory are not trivial, while the results shown in DeepVO \cite{DeepVO} were quite promising. This could possibly result from the intrinsic differences between the TUM RGB-D benchmark \cite{RGBD} we are using and the KITTI dataset \cite{KITTI} used in DeepVO. The videos collected in KITTI have smooth transitions between adjacent frames, while the transitions between frames in the TUM RGB-D benchmark are comparatively dramatic and often blurred, so predictions on our dataset is harder.
    
    \subsection{Transformer over Attention and LSTM}
    Possible reasons why our transformer-based model performs better than the attention-based model are as follows. First, our transformer-based model uses 7 more attention layers than our attention-based model, boosting its expressive power. The positional encoding layer in the transformer also augments the self-attention mechanism. Second, information extracted by attention layers must pass through a second LSTM layer in our attention-based model, while the transformer-based model removes the LSTM layer entirely and directly outputs a position estimate. As we have already observed LSTM-based VO models struggle to propagate information through time, it is possible that the LSTM layer undermines the effects of attention.

    \subsection{Future Work}
    The dataset that we are using contains only a few trajectories, so we could include other datasets for more training and validation data. The attention models that we developed do not incorporate a mask that prevents look-ahead to future image frames, so they are only capable of solving offline VO tasks. Naturally, we can add a mask that filters out all future information to adapt to online tasks. The hyperparameters of the transformer that we used are not tuned strategically using grid search or random search, so a more careful choice of hyperparameters such as number of attention heads, attention layers, and weight decay coefficients could possibly lead to improved performance. Currently we are using full attention layers on the whole sequence, and unrelated frames may add noise to the encoding. Thus we could use sparse attention like BigBird \cite{big_bird} to reduce noise and focus only on more relevant frames. In our models, we flatten the output of the CNN to generate image embeddings for the sequence models. Instead of doing so, we can replace the CNN with a vision transformer \cite{ViT} to attend to spatial information within each frame and generate image embeddings, and then pass these embeddings to our temporal transformer. This combination of spatial and temporal attention may offer significant improvements in model performance. 
    
\newpage
\bibliographystyle{unsrt}
\bibliography{refs}

\end{document}